\pdfoutput=1

\documentclass[11pt]{article}

\usepackage{ACL2023}

\usepackage{times}
\usepackage{latexsym}
\usepackage{booktabs}
\usepackage{hyperref}
\usepackage{graphicx}
\usepackage{multirow}
\usepackage{booktabs}

\usepackage[T1]{fontenc}

\usepackage[utf8]{inputenc}

\usepackage{microtype}

\usepackage{inconsolata}

\title{Mashee at SemEval-2024 Task 8: The Impact of Samples Quality on the Performance of In-Context Learning for Machine Text Classification}

\author{Areeg Fahad Rasheed \\
  College of Information Engineering \\
  Al-Nahrain University \\
  Baghdad, Iraq \\
  \texttt{areeg.fahad@coie-nahrain.edu.iq} \\\And
  M. Zarkoosh \\
  Software Engineering \\
  Baghdad, Iraq \\
  \texttt{m94zarkoosh@gmail.com} \\}

\begin{document}
\maketitle
\begin{abstract}
Within few-shot learning, in-context learning (ICL) has become a potential method for leveraging contextual information to improve model performance on small amounts of data or in resource-constrained environments where training models on large datasets is prohibitive. However, the quality of the selected sample in a few shots severely limits the usefulness of ICL. The primary goal of this paper is to enhance the performance of evaluation metrics for in-context learning by selecting high-quality samples in few-shot learning scenarios. We employ the chi-square test to identify high-quality samples and compare the results with those obtained using low-quality samples. Our findings demonstrate that utilizing high-quality samples leads to improved performance with respect to all evaluated metrics.

\end{abstract}


\section{Introduction}
The advent of large language models (LLMs) like GPT-3.5 has brought about transformative capabilities, seamlessly handling tasks like question answering, essay writing, and problem-solving \cite{aljanabi2023chatgpt, wu2023brief, rasheed2023arabic}. However, this technological advancement necessitates careful consideration of its associated challenges. Concerns regarding the potential impact on creativity and ethical implications, particularly concerning the generation of deepfakes \cite{tang2023science}, warrant careful attention \cite{projectmanagers}. Additionally, the limitations of LLMs, including the possibility of producing erroneous information, require rigorous evaluation and verification. The substantial energy consumption required for training LLMs on massive datasets raises environmental concerns, contributing to their carbon footprint. Moreover, plagiarism issues emerge as users may misuse the generated content, either inadvertently or intentionally \cite{hadi2023survey}. 

Various models have been introduced in recent years designed to distinguish text generated by humans from that created by machines\cite{mitchell2023detectgpt}. Examples include GPTZero\cite{gptzero}, AI Content Detector\cite{copyleaks}, and AI Content Detector by Writer \cite{writerai} among others. Some of these models are trained on specific datasets, while others are commercially available. Designing and implementing LLMs for classification tasks requires substantial resources and computational power, which are often only accessible to institutions and governments. Therefore, various optimization models, such as LoRA \cite{hu2021lora}, distillation\cite{hsieh2023distilling}, quantization\cite{dettmers2022llm}, and in-context learning \cite{liu2022few}, have been developed to reduce the resource requirements for LLM implementation. This paper focuses on In Context Learning (ICL) \cite{liu2022few}, which utilizes the capabilities of other models to enhance their ability to classify AI-generated text.

In Context Learning (ICL) is a Natural Language Processing (NLP) technique utilized to enable Large Language Models (LLMs) to learn new tasks based on minimal examples. This technique proves powerful in scenarios where training models on extensive datasets is impractical or when there are constraints on dataset availability for a specific task. ICL operates on the premise that humans can often acquire new tasks through analogy or by observing a few examples of task performance. It can be employed without any examples and is referred to as zero-shot learning. Alternatively, if the input includes one example, it is termed one-shot learning, and if it contains more than one, it is known as few-shot learning. This paper focuses on the application of few-shot learning within the context of ICL\cite{ahmed2022few, kang2023large}. 

In this study, our focus lies exclusively on few-shot learning. We present a methodology that leverages the chi-square statistic \cite{rasheed2023impact, lancaster2005chi} to select samples for few-shot learning and evaluate its impact on the performance of a machine-generated text classification model. We work on task A English language only  \cite{semeval2024task8}.

\section{Dataset}

The dataset employed for Task A comprises two main components. The first part, derived from human writing, was collected from diverse sources including WikiBidia, WikiHow, Reddit, ArXiv, and PeerRead. The second part consists of a machine-generated text produced by ChatGPT, Cohere, Dolly-v2, and BLOOMz\cite{muennighoff2023crosslingual}. For further details, please refer to the associated paper \cite{wang2023m4}. 

\section{Chi-square}
Chi-square is a statistical test used to assess the independence of two categorical variables. It calculates the difference between observed and expected frequencies of outcomes, and a larger chi-square value indicates a stronger rejection of independence. In text analysis, chi-square can be used to identify keywords that are more likely to occur in one category than another, making it useful for feature selection and text classification. We computed the chi-square values for each training sample and recorded the sample index with the highest and lowest chi-square values for both human-generated and machine-generated samples. Table I displays the index and corresponding chi-square values for each of these instances. We will use $X^2$ to refer to chi-square \cite{lancaster2005chi}.

\begin{table}[h]
  \centering
  \caption{Indices and chi-square values for highest/lowest in human-generated and machine-generated text}
  \begin{tabular}{ccc}
    \toprule
    \textbf{Name} & \textbf{Index \#} & \textbf{$X^2$ Value} \\
    \midrule
    Highest $X^2$ (Human)&70873& 1351.59\\
    Lowest $X^2$ (Human)&85726&1.21 \\
    Highest $X^2$ (Machine)&2426& 1154.27  \\
    Lowest $X^2$ (Machine)&29111& 0.8243 \\
    \bottomrule
  \end{tabular}
  \label{tab:simple}
\end{table}

\section{System overview}
\begin{figure*}
    \centering
    \includegraphics[width=0.8\linewidth]{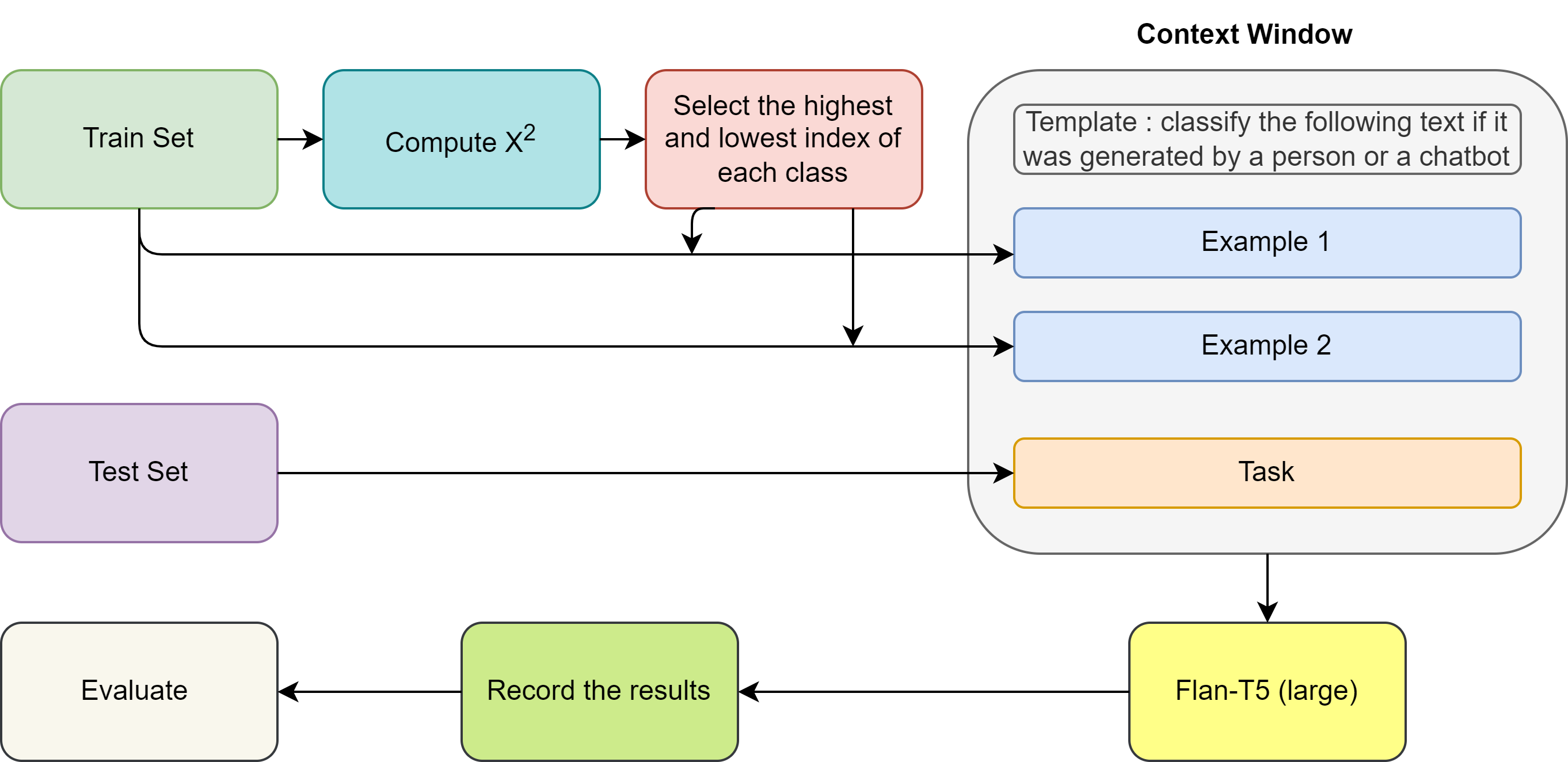}
    \caption{Proposed System Components}
    \label{fig:enter-label}
\end{figure*}
 The system architecture is illustrated in Figure 1. The process starts with feeding the entire training dataset to a chi-square computation, where the chi-square value for each sample is calculated. Subsequently, the indices of the samples with the highest and lowest chi-square values are selected for both human-generated and machine-generated datasets using information from Table I. Next, context learning is prepared. Initially, multiple templates were tested, and the one presented in Figure 1 yielded the best results. This template is then fed with two samples: the first being the machine-generated sample with the highest chi-square value, and the second being the human-generated sample with the highest chi-square value. Due to context window size limitations, only the first 5000 characters of each sample are incorporated. This is applied to training samples exceeding 5000 characters to ensure the context learning size is not exceeded. Finally, the test sample is fed into the context-learning process. The Flan-T5 model large version is used. The results are then recorded and evaluated. The dev/test sample size was truncated to 3000. We also evaluated the system using samples with the lowest chi-square values and doing the same process. 

\begin{table*}[h]
  \centering
  \begin{tabular}{cccccc}
    \toprule
    \textbf{Dataset}  & \textbf{Chi Type} & \textbf{Recall} & \textbf{Precision} & \textbf{F1-Score} & \textbf{Accuracy} \\
    \midrule
    \multirow{2}{*}{Dev set}   &Lowest & 46.92 & 46.90 & 46.84 &46.92\\
    & Highest & 53.76 & 53.76 & 53.74& 53.76 \\
    \cmidrule{2-6}
    \multirow{2}{*}{Test set} & Lowest & 55.04 & 55.07 & 55.03&55.27 \\
    &  Highest &58.68 & 58.81 & 58.81& \textbf{55.99} \\
    \bottomrule
  \end{tabular}
  \caption{Experiments results}
  \label{tab:row-spanning}
\end{table*}

 \section{Findings and Analysis}

We employed the Flan-T5 Large model for both the development and testing datasets. We selected samples from both human-generated and machine-generated sources, with each sample limited to 5000 characters to avoid exceeding the token size limit. A total of four experiments were conducted. The first experiment utilized samples with high chi-square values from the development set. The second experiment focused on samples with the smallest chi-square values from the development set. The third experiment involved samples with high chi-square values from the test set. Finally, the fourth experiment utilized samples with low chi-square values from the test set. Table II presents all achieved results.

Based on the results presented in Table II, we can discuss several key points. 

\begin{itemize}
    \item The results highlight the crucial role of sample quality in the performance of in-context learning. By leveraging the chi-squared metric and prioritizing samples with high values, we essentially provide the Flan-T5 model with examples rich in diverse features. This choice enables the Flan-T5 model to learn more effectively, drawing substantial insights from the samples. Consequently, the model becomes more familiar with the provided data, ultimately enhancing its performance. In contrast, selecting samples with lower quality leads to less optimal performance. This can be noticed for both the dev and test set. The main reason behind this is that words in the sample with high chi-square values contain the most distinctive features. This is because the chi-square test assigns high values to words that are frequent within a particular class but appear less frequently in other classes.Conversely, samples with lower chi-square values likely contain more random words that appear with similar frequency across all classes. In chi-square analysis, words that appear equally or approximately equally in each class receive lower scores.

    \item The classification of machine-generated text represents a novel frontier in machine learning, and the availability of datasets for this task is currently limited. The dataset used in this study was generated in 2023, marking it as a recent development and underscoring the lack of established benchmarks. Models that support in-context learning have not been trained extensively on such tasks, resulting in lower accuracy when applied. While examples with high-quality data can enhance model performance, it remain below the desired threshold. Hence, it is advisable to train the model directly on the dataset rather than relying on in-context learning. 
    
    \item We have utilized the Flan-T5 model; however, other models can be employed to evaluate the performance of text classification machinery. We suggest considering alternatives such as bard, Jurassic-1 Jumbo, and ChatGPT.

\end{itemize}

\section{Conclusion}
This work presents a system for classifying human-generated and machine-generated text. The system leverages the combined strengths of in-context learning and Chi-square analysis. Chi-square is employed to select high-quality samples from the trainin dataset for few-shot learning in the in-context learning. We implement Flan-T5 model large version for in-context learning. Evaluation using accuracy, recall, precision, and F1-score demonstrates that selecting high-quality samples improves system performance for both dev and test. Furthermore, the results indicate that relying solely on in-context learning for new tasks like machine-generated text detection yields relatively low performance.


\bibliography{custom}
\bibliographystyle{acl_natbib}

\end{document}